\documentclass[10pt,twocolumn,journal]{IEEEtran}
\ifCLASSINFOpdf
   \usepackage[pdftex]{graphicx}
\else
\fi
\usepackage{amsmath}
\usepackage{algorithm}
\usepackage[noend]{algorithmic}
\ifCLASSOPTIONcompsoc
 \usepackage[caption=false,font=normalsize,labelfont=sf,textfont=sf]{subfig}
\else
 \usepackage[caption=false,font=footnotesize]{subfig}
\fi

\newenvironment{myitemize}{\begin{list}{$\bullet$}
		{\setlength{\topsep}{1mm}
			\setlength{\itemsep}{0.25mm}
			\setlength{\parsep}{0.25mm}
			\setlength{\itemindent}{0mm}
			\setlength{\partopsep}{0mm}
			\setlength{\labelwidth}{15mm}
			\setlength{\leftmargin}{4mm}}}{\end{list}}

\usepackage{xcolor}

\begin{document}
%
\title{Collaborative Multi-Agent Video Fast-Forwarding}
%
%
%

\author{Shuyue Lan,~
        Zhilu Wang,~
        Ermin Wei,~
        Amit K. Roy-Chowdhury, ~\IEEEmembership{Fellow,~IEEE,}
       and Qi Zhu~
       }
\maketitle

%

\begin{abstract}

    Multi-agent applications have recently gained significant popularity. In many computer vision tasks, a network of agents, such as a team of robots with cameras, could work collaboratively to perceive the environment for efficient and accurate situation awareness. However, these agents often have limited computation, communication, and storage resources. Thus, reducing resource consumption while still providing an accurate perception of the environment becomes an important goal when deploying multi-agent systems. To achieve this goal, we identify and leverage the overlap among different camera views in multi-agent systems for reducing the processing, transmission and storage of redundant/unimportant video frames. Specifically, we have developed two collaborative multi-agent video fast-forwarding frameworks in distributed and centralized settings, respectively. In these frameworks, each individual agent can selectively process or skip video frames at adjustable paces based on multiple strategies via reinforcement learning.
    Multiple agents then collaboratively sense the environment via either 1) a consensus-based distributed framework called \textbf{DMVF} that periodically updates the fast-forwarding strategies of agents by establishing communication and consensus among connected neighbors, or 2) a centralized framework called \textbf{MFFNet} that utilizes a central controller to decide the fast-forwarding strategies for agents based on collected data.
   We demonstrate the efficacy and efficiency of our proposed frameworks on a real-world surveillance video dataset VideoWeb and a new simulated driving dataset CarlaSim, through extensive simulations and deployment on an embedded platform with TCP communication. 
   We show that compared with other approaches in the literature, our frameworks achieve better coverage of important frames, while significantly reducing the number of frames processed at each agent.

\end{abstract}


\section{Introduction}

\IEEEPARstart{W}{ith} the rapid advancement of camera sensors, a network of agents with cameras are increasingly being explored for tasks such as search and rescue, wide-area surveillance, and environmental monitoring, where the cameras may be built-in cameras in robots, cameras on drones, or fixed surveillance cameras. In these systems, multiple cameras can observe the same environment and generate videos from different angles, often with overlapping views, so that the fusion of all their perceptions may lead to better scene understanding. For many application tasks, this information fusion of large amount of data needs to be performed in real time or near real time. However, the agents often have limited computation, communication, storage, and energy resources
, which makes processing and transmitting all the video data quite challenging. This thus motivates the development of methods that can select an informative subset of the video frames to focus on. 

In the relevant literature, \emph{video summarization} and \emph{video fast-forwarding} both aim at generating a compact summary of the original video. In particular, video summarization methods often summarize videos in an offline manner, which needs an entire video available at hand before processing it~\cite{elhamifar2017online,gygli2015video,panda2017weakly,zhang2016video,zhao2014quasi}. Multi-view summarization methods that summarize videos from multiple cameras have also been proposed~\cite{fu2010multi,panda2016video,panda2017multi,elfeki2018multi,ou2015line}. However, as these methods process the entire videos and are often time-consuming, they are unsuitable for online and real-time applications. 
On the other hand, video fast-forwarding methods generate the video summary on the fly. Most of such methods adjust the playback speed of a video~\cite{cheng2009smartplayer,halperin2017egosampling,joshi2015real,petrovic2005adaptive,poleg2015egosampling,ramos2016fast,silva2016towards} while processing the entirety of it. One exception is our previous work FFNet~\cite{lan2018ffnet}, which performs video fast-forwarding for a single camera in an online manner and only processes a fraction of the video frames by automatically skipping unimportant frames via reinforcement learning. This shows promising results in reducing system computation and storage load. 
In this work, we build upon this approach and develop our solution for multi-agent video fast-forwarding systems.

\subsection{Solution Overview}

\begin{figure*}[t]
	\begin{center}
		\includegraphics[width=0.85\linewidth]{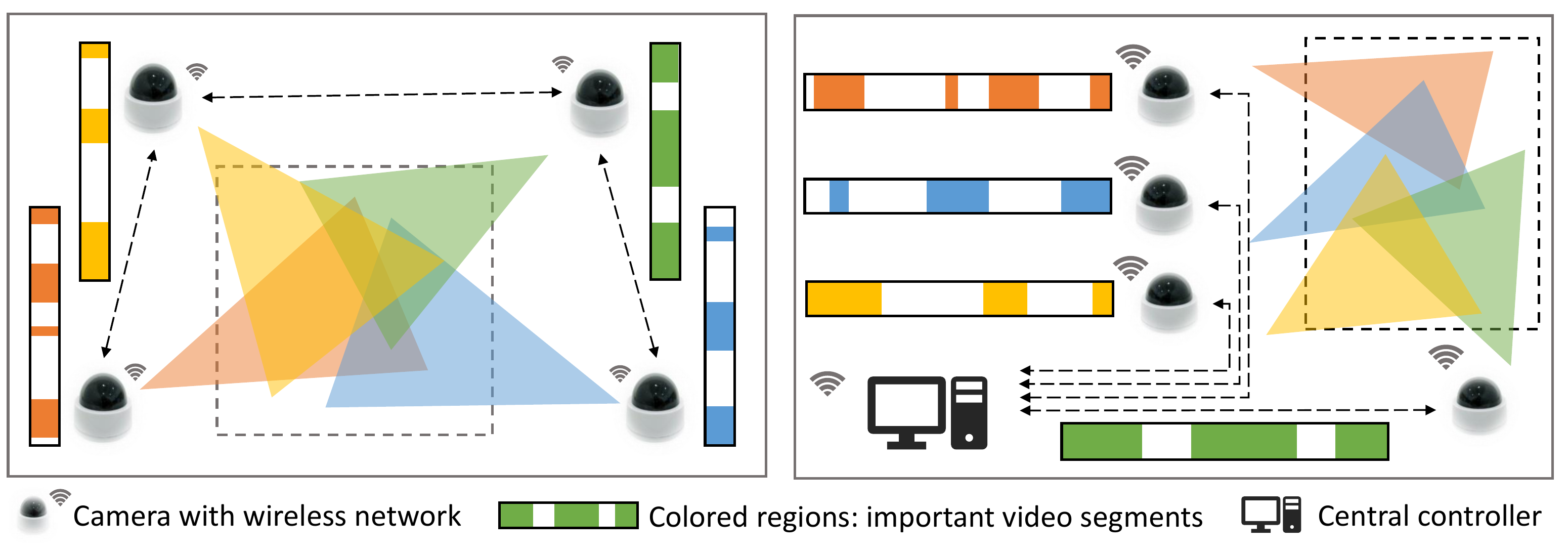}
	\end{center}
 \vspace{-6pt}
	\caption{\textbf{Illustration of collaborative multi-agent video fast-forwarding.} Multiple cameras at different agents are observing the same environment from different overlapping views. Each camera performs video fast-forwarding according to its current fast-forwarding strategy, which is decided either via communication and consensus among neighboring agents in a distributed manner (left) or by a central controller that analyzes the data from each agent (right). The colored regions within the bars represent the important video segments that each agent sees in its view. } 
	\label{overview}
 \vspace{-6pt}
\end{figure*}

Motivated by the observation that there is often significant overlap among videos captured by cameras from different angles in multi-agent systems, we pose the following question: \emph{Is it possible to leverage the overlapping among different views in multi-agent perception to collaboratively perform fast-forwarding that is efficient, causal, online, and results in an informative summary for the scene in real time?}

In this paper, we introduce two methods for multi-agent video fast-forwarding in  distributed and centralized settings, respectively. We target the scenarios where cameras at multiple agents observe the same environment from different angles. Each camera embeds a fast-forwarding agent with multiple strategies, i.e., it can skip the frames of its video input at different paces (e.g., slow, normal, or fast). During operation, each camera fast-forwards its own video stream based on a chosen pace and periodically updates its fast-forwarding strategies. 

For the distributed setting, part of our work has appeared in \cite{lan2020distributed}, named \textbf{DMVF}, which chooses and updates  fast-forwarding strategies by establishing communication and consensus among connected agents, as shown in the left figure of Fig.~\ref{overview}. Agents are connected by a predetermined undirected communication network\footnote{Note that some agents may not be able to communicate with each other due to practical factors such as the connection capacity of camera nodes, the physical distances between the nodes, the network bandwidth, etc.}, where each agent can communicate with a set of neighboring agents. At every adaptation period, each agent evaluates the importance of the selected frames from itself and those from its neighbors by comparing their similarities. Then a system-wide consensus algorithm is run among agents to reach an agreement on the importance score for every agent's view. Finally, based on the score ranking and the system requirement, each agent selects a fast-forward strategy for its next adaptation period.

For the centralized multi-agent video fast-forwarding setting, we have developed a new framework in this work, named \textbf{MFFNet}, which contains a central controller to decide the fast-forwarding strategies for each agent (the right part of Fig.~\ref{overview}). During operation, each camera fast-forwards its own video stream based on a chosen pace given by the central controller, and periodically sends selected frames (i.e., fast-forwarded clips) to the central controller. The central controller receives the selected frames from every agent and composes a more compact summary video for the scene. Moreover, based on the data at hand, the central controller infers the strategy/pace that should be adopted by each agent for the next period and sends such instruction back to the agents. Intuitively, an agent whose view currently contains more important frames than others should be slowed down for the next period to collect more frames; while agents whose views have significant overlaps with the slowed-down agents can be given a faster pace to reduce their processing and transmission load.

In both distributed and centralized settings, each agent only processes a very small portion of frames with fast-forwarding, which significantly reduces the computation load. The agents also do not require transmitting or storing their entire video streams (often only a fraction of them). From the system perspective, both the intra-view at each agent and the inter-view redundancy across different agents are reduced. Furthermore, the online and causal nature of our proposed approaches enables the users to begin fast-forwarding at any point when executing certain multi-agent perception tasks. Our approach is particularly useful for resource-constrained and time-critical systems such as multi-robot systems. 

The main contributions of this paper include the following.
\begin{itemize}
    \item We formulate the multi-agent video fast-forwarding problem as a collaborative multi-agent reinforcement learning problem. Each agent can fast-forward its video input without processing the entire video and be easily adapted to different fast-forwarding strategies/paces.
    \item Building upon our work in single-agent fast-forwarding (FFNet)~\cite{lan2018ffnet} and distributed multi-agent fast-forwarding (DMVF)~\cite{ lan2020distributed}, we develop a new centralized framework MFFNet for multi-agent fast-forwarding, which uses a central controller to orchestrate the fast-forwarding strategies of agents for achieving better scene coverage with reduced computation and communication load.
    \item We demonstrate the effectiveness of MFFNet on a challenging multi-view dataset, VideoWeb~\cite{denina2011videoweb}, achieving real-time speed on an embedded platform with TCP communication. We compare MFFNet with DMVF, FFNet, and a few other methods in the literature.
    \item  Moreover, for a more comprehensive comparison, we also include a newly generated multi-camera dataset for multi-agent video fast-forwarding, named CarlaSim, to further evaluate the various methods on moving platforms.
\end{itemize}

In particular, beyond our recent work~\cite{lan2020distributed}, this paper introduces the new development of 1) the MFFNet method, 2) the new CarlaSim dataset, and 3) the experimental results and analysis of MFFNet, as well as its comparison with DMVF, FFNet and other methods on VideoWeb and CarlaSim.

\subsection{Paper Organization}
This paper highlights our new contributions in MFFNet and also  introduces our prior work in FFNet and DMVF, providing a holistic view of our solution in video fast-forwarding. More specifically, FFNet is a single-agent video fast-forwarding method that we developed based on reinforcement learning, and we build a multi-strategy video fast-forwarding agent upon FFNet. Both DMVF and MFFNet use this multi-strategy fast-forwarding agent on their camera nodes -- DMVF uses a distributed framework to decide the strategies each agent should use, while MFFNet uses a centralized framework to do so. Both methods are efficient and effective on collaborative video fast-forwarding for a network of resource-limited agents.

In the rest of the paper, we first present a review of relevant literature in Sec.~\ref{sec:related}. This is followed by a review of our work in developing FFNet for single-agent video fast-forwarding in Sec.~\ref{sec:ffnet}, along with the development of a multi-strategy video fast-forwarding agent. In Sec.~\ref{sec:dmvf} and Sec.~\ref{sec:mffnet}, we present our solutions to the multi-agent video fast-forwarding problem for distributed and centralized settings, i.e., DMVF and MFFNet, respectively. Experimental results in real-life data are presented in Sec.~\ref{sec:exp}.

\section{Related Work}\label{sec:related}

\subsection{Video Summarization and Video Fast-forwarding}

The objective of video summarization is to take an entire video as input and output a compact subset of frames that can describe the important content of the original video. Many single-view video summarization methods are developed with unsupervised learning~\cite{elhamifar2012see,gygli2014creating,Top2014,elhamifar2019unsupervised} and supervised learning techniques based on video-summary labels~\cite{gygli2015video, panda2017weakly, zhang2016video,gong2014diverse,wu2019adaframe, rochan2019video}. There are methods proposed specifically for summarizing crawled web images/videos~\cite{khosla2013large,Joint2014,song2015tvsum,panda2017collaborative}  and photo albums~\cite{sigurdsson2016learning}, and online methods developed using submodular optimization~\cite{elhamifar2017online}, Gaussian mixture model~\cite{ou2014low}, and online dictionary learning~\cite{zhao2014quasi}. Beyond single-view, the multi-view video summarization problem has been addressed by random walk over spatio-temporal graphs~\cite{fu2010multi}, joint embedding and sparse optimization~\cite{panda2016video,panda2017multi}, DPP (Determinantal Point Processes)~\cite{elfeki2018multi}, and a two-stage system with online single-view summarization and distributed view selection~\cite{ou2015line}. Different from these methods, our approaches do not process all the frames, which significantly reduces computation and communication load, and they collaboratively fast-forward multi-view videos, further improving the efficiency and coverage. 

Video fast-forwarding methods are used for skipping uninteresting/unimportant parts of the video. Commercial video players often offer the users with manual control on the playback speed, such as Apple QuickTime player with 2x, 5x, and 10x speed fast-forward. 
In the literature, the playback speed can be automatically adjusted based on the similarity of each candidate clip to a query clip~\cite{petrovic2005adaptive} and the motion activity patterns in videos~\cite{cheng2009smartplayer,peker2003extended,peker2001constant}.
Besides playback speed adjustment, some works develop the fast-forwarding policy based on mutual information between frames~\cite{jiang2010new,jiang2011smart}, shortest path distance over the semantic graph built from frames~\cite{ramos2016fast,silva2016towards}, and visual and textual features~\cite{ramos2020straight}. 
Hyperlapse is also widely studied for fast-forwarding videos aiming at speed-up and smoothing~\cite{poleg2015egosampling,halperin2017egosampling,joshi2015real}. 
Different from these approaches that are for single videos, our work focuses on multi-agent video fast-forwarding methods that collaboratively fast-forward videos from different views.

\subsection{Reinforcement Learning}
Deep reinforcement learning has been widely used in many computer vision tasks and achieved promising performance, such as in action detection~\cite{yeung2016end}, object detection~\cite{Mathe_2016_CVPR}, image captioning~\cite{Ren_2017_CVPR}, pose estimation~\cite{Krull_2017_CVPR}, visual tracking~\cite{Yun_2017_CVPR} and query-conditioned video summarization~\cite{zhang2019deep}. 
There are also approaches applying reinforcement learning to the multi-agent domain, i.e., multi-agent reinforcement learning (MARL) (see a detailed review in~\cite{bu2008comprehensive}).    
Some recent works have used MARL to address computer vision tasks, such as joint object search~\cite{kong2017collaborative}, multi-object tracking~\cite{ren2018collaborative}, and frame sampling for video recognition~\cite{wu2019multi}. 
There are also works on building learnable communication protocols for collaborative multi-agent deep reinforcement learning~\cite{sukhbaatar2016learning,foerster2016learning}. Our earlier work FFNet conducts single video fast-forwarding via reinforcement learning~\cite{lan2018ffnet}, based on which we further develop two approaches for multi-agent video fast-forwarding in centralized and distributed settings.

\subsection{Multi-agent System Optimization}

A fundamental problem in distributed multi-agent systems is the minimization of a sum of local objective functions while maintaining agreement over the decision variable, often referred to as consensus optimization. Seminal work in~\cite{tsitsiklis1984problems} proposes a distributed consensus protocol for achieving agreement in a multi-agent setting by iteratively taking a weighted average with local neighbors. The work in~\cite{nedic2009distributed} presents a distributed gradient descent (DGD) method, where each agent iteratively updates its local estimate of the decision variable by executing a local gradient descent step and a consensus step.  Follow-up works~\cite{NOPConstrained,MateiBaras, nedic2011asynchronous} extend this method to other settings, including stochastic networks, constrained problems, and noisy environments.  
More recently, EXTRA~\cite{shi2015extra}, which takes a careful combination of gradient and consensus steps, is proposed to improve convergence speed and is shown to achieve linear convergence with constant step size. In computer vision, consensus-based methods are used applications such as human post estimation~\cite{lifshitz2016human}, background subtraction~\cite{wang2006background}, and multi-target tracking~\cite{kamal2015distributed}, etc. To the best of our knowledge, the DMVF framework (more details on~\cite{lan2020distributed}) we developed is the first distributed consensus-based framework to address multi-agent video fast-forwarding. In this paper, we further develop a centralized framework MFFNet that facilitates a central controller to adjust the fast-forwarding strategy for multi-agent video fast-forwarding.

\section{Single-agent Video Fast-forwarding} \label{sec:ffnet}
\subsection{Review of FFNet}

FFNet~\cite{lan2018ffnet} uses a Markov decision process (MDP) to formulate the video fast-forwarding problem and solves it using reinforcement learning, i.e., with a Q-learning agent that learns a policy to skip unimportant frames and present the important ones for further processing. Given the current frame, FFNet decides the number of frames to skip next. The MDP formulation of FFNet is defined as follows:
\begin{myitemize}
    \item \textbf{State:} A state $s_k$ describes the environment at time step $k$. It is defined as the feature vector of the current frame.
    
    \item \textbf{Action:} An action $a_k$ is performed by the system at step $k$ and devotes to an update of the state. The action set includes the possible numbers of frames to skip.
    
    \item \textbf{Reward:} An immediate reward $r_k = r(s_k, a_k, s_{k+1})$ is received by the system at time step $k$ as
    \begin{equation} 
    	r_k =   - SP_k +HR_k.
    \end{equation}
    It consists of the ``skip'' penalty (SP) and the ``hit'' reward (HR). $SP_k$ defines the penalty for skipping action in the interval $t_k$ at step $k$:
    \begin{equation}
	SP_k =  \frac{\sum_ {i \in t_k}^{ }\textbf{1}(l(i) = 1)}{T} - \beta \frac{\sum_{i \in t_k}^{ } \textbf{1}(l(i) = 0)}{T},
    \end{equation}
    where $\textbf{1}(\cdot)$ is an indicator function that equals to 1 if the condition holds. T is the largest number of frames we may skip. $\beta \in [0,1]$ is a trade-off factor between the penalty for skipping important frames and the reward for skipping unimportant frames.
    $HR_k$ defines the reward for jumping to an important frame or a position near an important frame and is computed as 
    \begin{equation}
	HR_k =  \sum_{i=z-w}^{z+w} \textbf{1}(l(i)=1) \cdot f_i(z),
    \end{equation}
    where $f_i(z)$ extends the one-frame label at frame i to a Gaussian distribution in a neighboring time window $w$, i.e., $z \in [i-w, i+w]$. 
    
    \item \textbf{Policy:} With the definition of states, actions, and rewards, a skipping policy $\pi$ is learned for selecting the action that maximizes the expected accumulated reward $R$:
    \begin{equation}
	\pi(s_k) = arg \max_{a} E[R|s_k, a, \pi],
    \end{equation}
    where the accumulated reward $R$ is computed as 
        \begin{equation} 
        	R = \sum_{k}^{} \gamma^{k-1}r_k = \sum_{k}^{} \gamma^{k-1}r(s_k, a_k, s_{k+1}),
        \end{equation}
    where $\gamma \in [0, 1] $ denotes the discount factor for the rewards in the future.  
\end{myitemize}

With Q-learning, the value of $E[R|s,a,\pi]$ is evaluated as $Q(s,a)$. The optimal value $Q^*(s_k,a_k)$ can be calculated by the Bellman equation in a recursive fashion:
\begin{equation} \label{bellman}
	Q^*(s_k,a_k) = r_k + \gamma \max_{a_{k+1}}Q^*(s_{k+1}, a_{k+1}).
\end{equation}

\begin{figure}[t]
	\begin{center}
		\includegraphics[width=0.85\linewidth]{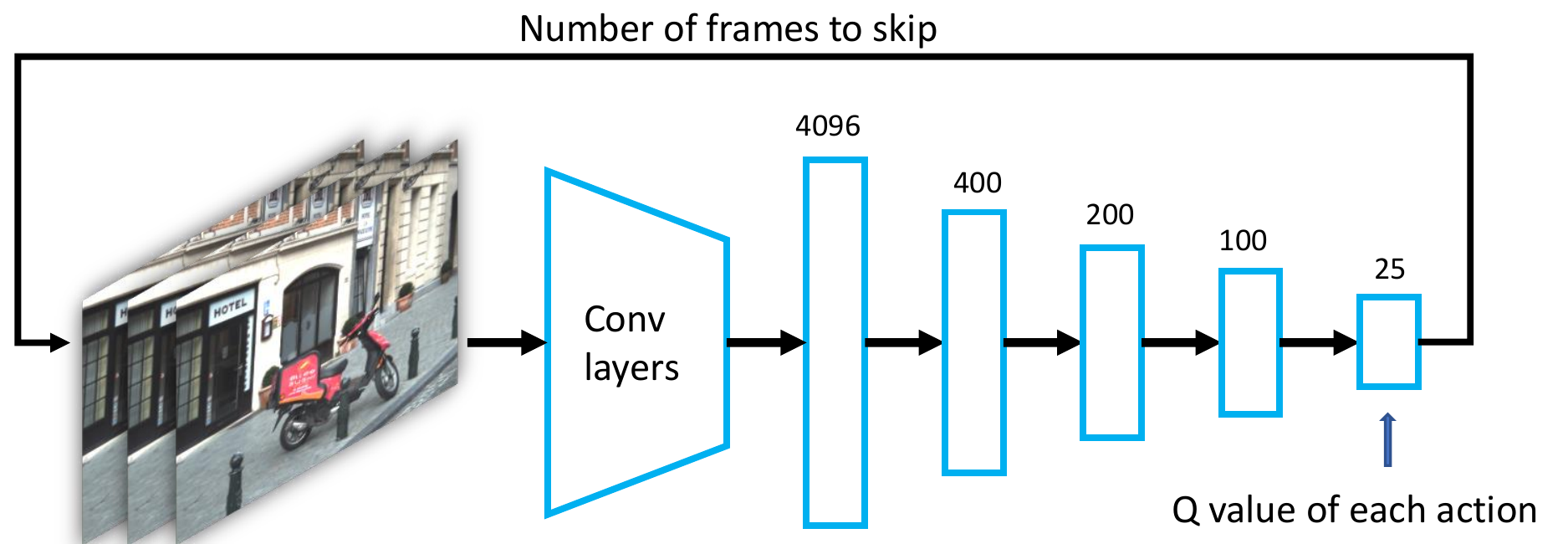}
	\end{center}
	\caption{\textbf{The model structure of FFNet.} It takes a frame in an incoming video stream as an input for the deep neural network and outputs the number of frames to skip.}
	\label{ffnet}
 \vspace{-6pt}
\end{figure}

The model of FFNet is shown in Fig.~\ref{ffnet}. When training this model, the mean squared error between the target Q-value and the output of MLP is used as the loss function. $\epsilon$-greedy strategy is utilized to better explore the state space, which picks a random action with probability $\epsilon$ and the action that has $Q^*(s,a)$ with probability 1-$\epsilon$. 

\subsection{Multi-strategy Fast-forwarding Agent} \label{sec:agent}

To fit into the multi-agent video fast-forwarding scenario, on each camera that captures a view of the scene, we leverage a multi-strategy fast-forwarding agent that can adaptively fast-forward the incoming videos with different paces. Similar to \cite{lan2020distributed}, the FFNet is derived into three different strategies/paces for fast-forwarding: normal-pace, slow-pace, and fast-pace. Note that our approach can be easily extended to consider other numbers of strategies/paces.

\medskip \noindent \textbf{Normal-pace Strategy.} The normal-pace strategy adopts the same immediate reward design as FFNet:
\begin{equation}
r_k(normal) =   - SP_k +HR_k.
\end{equation}
As our normal-pace strategy, we use an action space of size 25, i.e., skipping from 1 to 25 frames.

\medskip \noindent \textbf{Slow-pace Strategy.}
The slow-pace strategy aims at skipping fewer frames and thus retaining more frames in the selected buffer, possibly including more numbers of important frames. To meet this goal, we modify the immediate reward in FFNet at time step $k$ as 
\begin{equation}
r_k(slow) =   (- SP_k +HR_k) \times (1 - \frac{sigmoid(a_k)}{2}).
\end{equation}
Intuitively, if the agent skips a larger step, it will receive a smaller immediate reward. We also change the action space to 15 to prevent the agent from skipping too much.

\medskip \noindent \textbf{Fast-pace Strategy.}
The goal of the fast-pace strategy is to skip more unimportant frames for more efficient processing and transmission. Thus, we modify the immediate reward at time step $k$ as
\begin{equation}
r_k(fast) =   (- SP_k +HR_k) \times (1 + \frac{sigmoid(a_k)}{2}).
\end{equation}
This reward definition ensures that the agent will get a larger immediate reward if it skips a larger step. The action space is set to 35 to allow the agents to skip larger steps.
For each agent, it can flexibly switch among these strategies to adaptively fast-forward its own videos.

 \section{DMVF: Distributed Multi-agent Video Fast-forwading} \label{sec:dmvf}

\subsection{Overview}

\begin{figure*}[t]
	\begin{center}
		\includegraphics[width=0.85\linewidth]{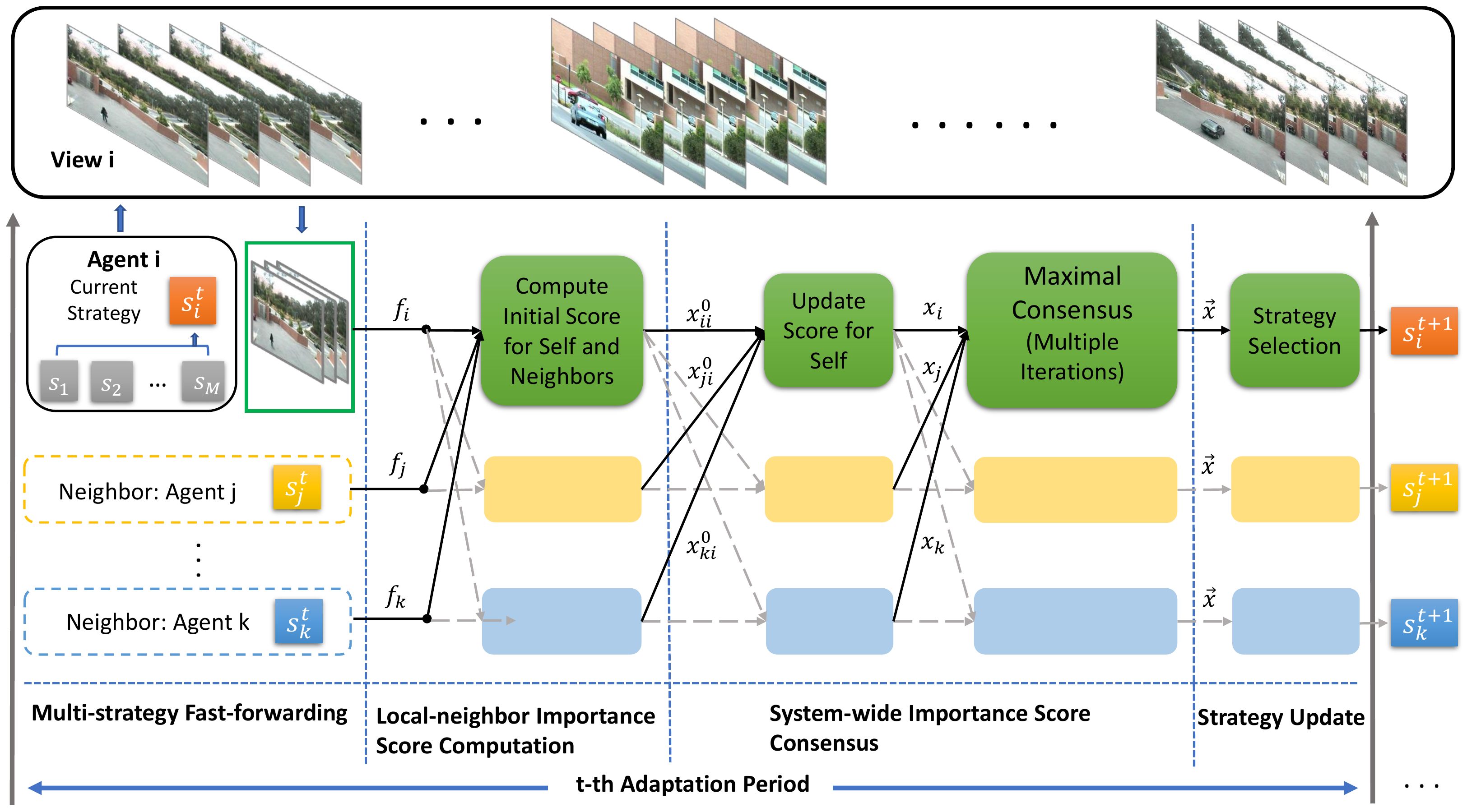}
	\end{center}
\vspace{-6pt}
	\caption{\textbf{The workflow of DMVF}. At every adaptation period $t$, each agent $i$ first fast-forwards its video input with current strategy $s_i^t$ and selects a set of frames $f_i$. It then receives neighbor agents' selected frames (e.g., $f_j$ and $f_k$) and computes an initial importance score for itself and its neighbors. Afterwards, agent $i$ refines and finalizes the importance score with other agents via a system-wide maximal consensus algorithm. Based on this importance score vector $\vec{x}$, agent $i$ chooses its strategy for the next period $s_i^{t+1}$ (so does every other agent).}
	\label{framework_dmvf}
\vspace{-6pt}
\end{figure*}

In this section, we review our approach for addressing the multi-agent video fast-forwarding problem by adapting the skipping strategy of each agent in an efficient, online, and distributed manner, named DMVF (more details in \cite{lan2020distributed}). Fig.~\ref{framework_dmvf} shows the workflow design of our framework (take one agent $i$ for illustration). Given the incoming multi-view video streams $V = \left\{v_1, \cdots, v_N\right\}$ captured at different agents, our goal is to generate a final summary $F = \left\{f_1,\cdots, f_N\right\}$ for the scene while reducing the computation, communication, and storage load. 

In our framework, the fast-forwarding agent of each view is modeled as a reinforcement learning agent with multiple available strategies $S=\left\{s_m, m = 1, \cdots , M\right\}$. During operation, at every adaptation period $t$ (with the period length as $T$), each agent $i$ fast-forwards its own video stream with a current strategy $s_i^t \in S$ and selects a subset of frames $f_i$. Note that the frames being skipped are not processed, transmitted, or saved.   
Agent $i$ then communicates with its neighbors and receives their selected frames, e.g., $f_j$ and $f_k$ as shown in the figure. Based on such information, agent $i$ computes an initial importance score for itself and its neighbors.
Afterward, agent $i$ refines its initial score together with other agents in the system via a system-wide consensus algorithm, including first an update of its own score and then multiple iterations to reach system-wide consensus. Note that during the consensus process, only scores are transmitted (not selected frames).
After running the consensus algorithm, each agent will have the same copy of the final importance scores for their selected frames in the current period, defined as $\vec{x} = [x_1, x_2, ..., x_N]$. Agent $i$ then chooses its fast-forwarding strategy for the next period $s_i^{t+1}$ based on the rank of its importance score $x_i$. The notations are highlighted in Table~\ref{tab:notation}.

\begin{table}[t]
\begin{center}
\begin{tabular}{c|l}
\hline
$M$  & number of available fast-forwarding strategies\\
\hline
$N$  & number of camera views / agents\\
\hline
$V$ & the set of $N$ views $\left\{v_i\right\}$, $i \in [1, N]$  \\
\hline
$S$   & the set of available strategies $\left\{s^m\right\}$, $m \in [1, M]$ \\
\hline
$s_i^t$  & strategy being used in agent $i$ at adaptation step $t$\\
\hline
$s_i^{t+1}$   & strategy for agent $i$ in the next adaptation step $t+1$\\
\hline
$F$  & summary of the scene: $\left\{f_1,\cdots, f_N\right\}$ \\
\hline
$\vec{x}$ & importance score vector after consensus \\
\hline
$T$   & period of strategy update \\
\hline
\end{tabular}
\end{center}
\caption{Notations used in DMVF.}
\vspace{-12pt}
\label{tab:notation}
\end{table}

\subsection{Local-neighbor Importance Score Computation}
\label{sec:local_score}

In this step, for every agent $i$, we compute an initial importance score for itself and its neighbor by comparing the similarities between their selected frames.
First, we evaluate the similarity between two frames $x$ and $y$ by computing the exponential of the scaled negative L2-norm of the feature representations of the two frames, as defined in the following equation:
\begin{equation} \label{eqn:sim}
    sim(x, y) = e^{-\alpha||x-y||_2},
\end{equation}
where $\alpha$ is used to scale the L2-norm to restrict the similarity value to a satisfactory range ($\alpha = 0.05$ in our experiment). 

The similarity of agent $j$ to $i$ is then defined as 
\begin{equation}
    sim\_agent (v_i, v_j) = \frac{1}{|v_j|} \sum_{s=1}^{|v_j|} \max_{1\leq a \leq |v_i|} sim(p_s(v_j), p_a(v_i)),
\end{equation}
where $|v_j|$ denotes the number of selected frames from agent $j$ and $p_s(v_j)$ denotes a selected frame $s$ from agent $j$. The similarity for frame $p_s(v_j)$ to agent $i$ is the maximum among the similarities between $p_s(v_j)$ and frames of agent $v_i$. Then the agent-to-agent similarity of agent $j$ to agent $i$ is the average frame similarity. 

We define the communication connections among agents as an undirected graph $G=(V,E)$. With this definition, we compute the importance score of agent $j$ estimated by agent $i$ as 
\begin{equation}\label{equ:xij0}
    x_{ij}^0 = \begin{cases}
        \frac{1}{|V_i|-1} \sum_{v_k \in V_i, k \neq j} sim\_agent (v_j, v_k) & 
        \\ \qquad \qquad \qquad \text{if } i=j \text{ or } (i,j) \in E  
        \\ 0  & \text{o.w.} 
\end{cases}
\end{equation}
where $V_i= \left\{v_k | (i,k) \in E\right\} \bigcup \left\{v_i\right\}$, is the set of the neighbors of agent $i$ and itself. $|V_i|$ represents the number of agents in $V_i$. This initial important score will then be refined via a consensus process as described in the following section. 

\begin{figure*}[t]
	\begin{center}
		\includegraphics[width=0.85\linewidth]{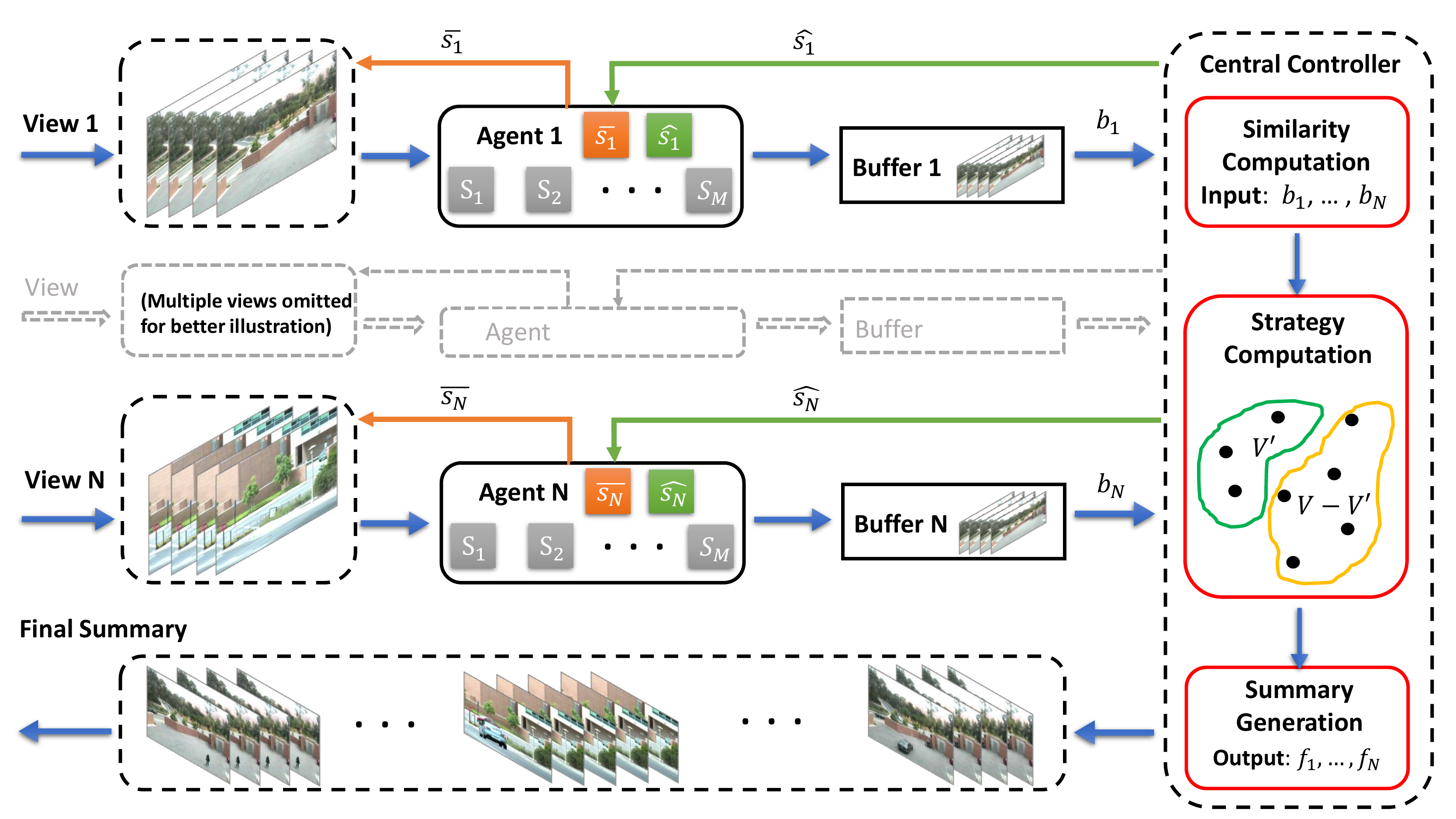}
	\end{center}
 \vspace{-6pt}
	\caption{\textbf{The workflow overview of MFFNet}. Each camera view is associated with an adaptive fast-forwarding agent that supports multiple fast-forwarding strategies/paces. During every period of operation, each agent $n$ uses current strategy $\Bar{s_n}$ to fast-forward its video input and saves selected frames in its buffer. At the end of the period, every agent sends the selected frames in its buffer to the central controller. The central controller computes the similarity among the frames from different agents, and based on it, chooses the strategy $\hat{s_n}$ for each agent $n$ in the next period and generates a more compact summary from their selected frames.}
	\label{framework}
 \vspace{-6pt}
\end{figure*}

\subsection{System-wide Importance Score Consensus} \label{sec:consensus}

To refine the initial importance score and reach an agreement across all agents on the relative importance of their frames, we mainly use a maximal consensus algorithm in our framework. We have also explored multiple variants of our framework with different consensus methods in \cite{lan2020distributed}. 

There are three steps in our maximal consensus algorithm. First, each agent communicates with its neighbors and sends its initial importance scores for each of them. At the end of this step, agent $i$ will have the initial scores of itself from its own computation and from the evaluation by its neighbors (i.e. \{$x_{ji}^0$\}, $j \in V_i$). Then, in the second step, agent $i$ updates its score as %
    \begin{equation}\label{equ:importantscore}
        x_i = \frac{\sum_{j \in V_i} \frac{1}{n_j} x_{ji}^0}{\sum_{j \in V_i} \frac{1}{n_j}},
    \end{equation}
which means that the importance score of agent $i$ is updated as the weighted average of the initial importance scores evaluated by itself and its neighbors. Then an importance score vector $\vec{x}_i$ for all agents is constructed by agent $i$, with only the $i$-${th}$ element set to $x_i$ and all others set to zero. 
In the third step, all agents will run a maximal consensus algorithm over the importance score vector. This algorithm only requires the number of consensus steps to be the diameter of the graph $G$ to reach an agreement (the convergence is guaranteed). In the end, every agent will have the same copy of the importance score vector for all agents, i.e., $\vec{x}_i =\vec{x} = [x_1, x_2, ..., x_N]$.

\subsection{Strategy Selection}
\label{sec:strategy_selection}

Based on the final importance scores in $\vec{x}$, the agents with higher scores could be assigned with a slower strategy for the next period, while the agents with lower scores could be faster.
Given the system requirement, the portions of different strategies are pre-defined, which means there should be a fixed number of agents under each strategy after every update.

\section{MFFNet: Centralized Multi-agent Video Fast-forwading} \label{sec:mffnet}

\subsection{Overview} \label{sec:overview}

In this section, we present a new method to address the multi-agent video fast-forwarding problem by utilizing a central controller to analyze the data from each agent and adapt the fast-forwarding strategies of agents in an efficient online manner, named MFFNet. Fig.~\ref{framework} shows the workflow design of our framework. Given the incoming multi-view video streams $V = \left\{v_1, \cdots, v_N\right\}$ captured at different agents, the goal of MFFNet is to generate a final summary $F = \left\{f_1,\cdots, f_N\right\}$ for the scene while reducing the computation, communication, and storage load.

The fast-forwarding agent of each camera view is modeled as a reinforcement learning agent with multiple available strategies $\left\{s_m, m = 1, \cdots , M\right\}$. During operation, each agent $n$ fast-forwards its own video stream with a current strategy $\Bar{s}_n$ and keeps the selected frames in its buffer $B_n$. The frames being skipped are not processed, transmitted, or saved. After a period of time $T$, each agent sends the selected frames in its buffer to the central controller. The central controller receives selected frames of the last period from all agents and computes their similarity. Based on the similarity computation, the controller chooses the strategy $\hat{s}_n$ for each agent $n$ in the next period and notifies them immediately. Such computation and decision are very fast and only performed once every period. The central controller also generates a more compact summary of the selected frames and stores them. The notations are highlighted in Tab.~\ref{tab:notation_2}.

\begin{table}
\begin{center}
\begin{tabular}{c|l}
\hline
$M$  & number of available fast-forwarding strategies\\
\hline
$N$  & number of camera views / agents\\
\hline
$v_n$ & the video of view $n$, $n \in [1,N]$.\\
\hline
$V$ & the set of $N$ views \\
\hline
$V'$ & the subset of $V$ containing selected main views\\
\hline
$s_m$   & available strategy $m$, $m \in [1, M]$ \\
\hline
$\Bar{s_n}$  & strategy being used in agent $n$\\
\hline
$\Hat{s_n}$   & strategy for agent $n$ in the next period\\
\hline
$\left\{A_n\right\}$    & set of fast-forwarding agents, $\left\{A_1, \cdots, A_N\right\}$ \\
\hline
$\left\{B_n\right\}$    & set of buffers, $\left\{B_1, \cdots, B_N\right\}$\\
\hline
$\left\{b_n\right\}$    & set of data received by controller, $\left\{b_1, \cdots, b_N\right\}$\\
\hline
$F$  & summary of the scene: $\left\{f_1,\cdots, f_N\right\}$ \\
\hline
$T$   & period of strategy update \\
\hline
$\rho$ & the threshold for matching frames\\
\hline
\end{tabular}
\end{center}
\caption{Notations used in MFFNet.}
\vspace{-12pt}
\label{tab:notation_2}
\end{table}

\subsection{Central Controller} \label{sec:controller}
The responsibility of the central controller is to decide the pace for each agent and generate a more compact summary of the scene. At every period $T$, it receives the selected frames $\left\{b_1, \cdots, b_N\right\}$ from all agents. With those data, it first computes similarity among frames from different agents. Based on the similarity, the central controller decides the new strategies $\left\{\hat{s}_1, \cdots, \hat{s}_n \right\}$ for all agents and sends them back. Meanwhile, the controller further reduces redundancy by generating a compact summary  $F = \left\{f_1,\cdots, f_n\right\}$. The central controller consists of three modules: similarity computation, strategy computation, and summary generation.

\medskip \noindent \textbf{Similarity Computation.}
From each agent $n$, the central controller receives a set of frames $b_n$ per period. In this module, the similarity between two frames is defined in Eqn.~\eqref{eqn:sim} in Sec.~\ref{sec:dmvf}. A threshold $\rho$ is used to match frames. If the similarity of two frames is greater than $\rho$, we consider them as a match. In order to further compute the strategies for each agent, we define a function named match count $M( \cdot , \cdot)$, which matches frames from two sources and returns the number of matching frames, as shown below: 
\begin{equation} 
    \label{equ:match_count}
    M(u, v) = \sum_{x \in u} I(\max_{y \in v}(sim(x,y)) > \rho),
\end{equation}
where $I(\cdot)$ is  an indicator function that equals 1 if the
condition holds.

\medskip \noindent \textbf{Strategy Computation.}
The goal of the strategy computation module is to infer the strategies for all agents in the next period, i.e., $\left\{\hat{s_1}, \cdots, \Hat{s_n} \right\}$. Intuitively, if a view contains a larger number of important frames, it should receive more attention and should not be skipped too much. Following this idea, we formulate the strategy computation problem as an optimization problem for selecting a subset of views $V'$ as the main views from $V$ to better represent the whole scene. The set of main views is selected by

\begin{equation}\label{equ:MainViewScore}
    V' = arg \max_{\Bar{V}} \frac{\sum_{i \in V-\Bar{V}} M(b_i, \bigcup_{j \in \Bar{V}} b_j )}{\sum_{j \in \Bar{V} }len(b_j)},
\end{equation}
where $b_i$ and $b_j$ are the frames sent back by the fast-forward agents $i$ and $j$. $len(\cdot)$ represents the number of frames in a fast-forwarded segment. The set of main views is selected as the subset of views that can cover the most of other views. To avoid the effect of the main view size, we divide the sum of match counts of other views by the total number of frames in the main view set. The detailed algorithm for main view set selection is shown in Algorithm~\ref{alg1}. Here,
$sz$ denotes the total number of frames in the main views. $score$ is the main view score of the subset $\Bar{V}$, as in Eqn.~\eqref{equ:MainViewScore}.

\begin{algorithm}[t]
	\caption{Main View Set Selection Algorithm}
	\label{alg1}
	\begin{algorithmic}[1]
		\STATE \textbf{Input:} a set of data received by the controller, $\left\{b_1, \cdots, b_N\right\}$, the similarity threshold $\rho$
		\STATE \textbf{Output:} A set of selected main views $V'$
		\STATE Initialize the similarity array $Similarity$
		\FOR{$i = 1$ to $N$}
		\FOR{$j = 1$ to $N$, $j\neq i$}
		\FOR{$k = 1$ to $Size(b_i)$}
		\FOR{$l = 1$ to $Size(b_j)$}
		\STATE $Similarity[i,j,k,l] = sim(b_i[k], b_j[l])$
		\ENDFOR
		\ENDFOR
		\ENDFOR
		\ENDFOR
		\STATE $MaxScore = 0$
		\FOR{$\delta = 1$ to $(2^N-2)$}
		\STATE $\Bar{V}=\{\}$, $sz = 0$, $score=0$
		\FOR{$i = 1$ to $N$}
        \IF{the $i$-th bit of $\delta$ is $1$}
		\STATE $\Bar{V} \leftarrow \Bar{V}\cup \{i\}$
		\STATE $sz \leftarrow sz + Size(b_i)$
		\ENDIF
		\ENDFOR
		\FOR{$i = 1$ to $N$, $i\notin \Bar{V}$}
		\FOR{$k = 1$ to $Size(b_i)$}
		\STATE $match = 0$ 
		\FOR{$j \in \Bar{V}$}
		\FOR{$l = 1$ to $Size(b_j)$}
		\IF{$sim(b_i[k], b_j[l]) > \rho$}
		\STATE $match \leftarrow 1$
		\ENDIF
		\ENDFOR
		\ENDFOR
		\STATE $score \leftarrow score + match$
		\ENDFOR
		\ENDFOR
		\STATE $score \leftarrow score / sz$
		\IF{$score > MaxScore$}
		\STATE $V'\leftarrow \Bar{V}$, $MaxScore\leftarrow score$
		\ENDIF
		\ENDFOR
	\end{algorithmic}	
\end{algorithm}

For the views in the main view set $V'$, they can cover more content than other views and have more important information. Thus, we use the slow fast-forwarding strategy for each of them. For any other views in $V-V'$, they can be covered significantly by the main views. Therefore, we expect them to fast-forward at a faster speed. More specifically, we decide their strategies by their matching percentage to the main view set. The matching percentage of view $n$ is computed as $mp(n) = M(b_n, \bigcup_{j \in \Bar{V}} b_j)/len(b_n)$. If the matching percentage of a view is smaller than a threshold $\tau$, it will be instructed to maintain the normal pace; otherwise, it will be instructed to use the fast strategy, as below: 
\begin{equation}
    \Hat{s}_n = \left\{
\begin{aligned}
&slow, &if \quad n \in V', \\
&normal, &if \quad n \in {V-V'}, mp(n) < \tau, \\
&fast, &if \quad n \in {V -V'}, mp(n) > \tau .
\end{aligned}
\right.
\end{equation}
 
 \medskip \noindent \textbf{Data Buffer and Strategy Update.}
For each agent $n$, there is a data buffer $B_n$ for storing the selected frames. At every time period $T$, the agent will send those frames in the buffer to the central controller. The agent will also receive a new strategy/pace instruction from the central controller and adapt it accordingly in the next period.
 
\medskip \noindent \textbf{Summary Generation.} After matching the frames among views and choosing the main view, the next step is to generate a more compact summary for the scene. We use the following policy for further reducing redundancy: 1) for the set of main views $V'$, we keep all the frames from its buffer in the summary, and 2) for the other views, we remove the frames that are matched with the main views (i.e., similar to some frames in the main views) and only keep the remaining ones in the summary. Please note that when generating the summary, we restrict the reduction of frames within a certain time window. That is, if two frames are similar with respect to the similarity threshold $\rho$ and are close to each other in time, we consider them as a match and drop it. Finally, similarly to~\cite{lan2018ffnet}, we also include some neighboring frames of the selected frames in the summary (with selected ones as the window centroids). All these summary frames are denoted as
$\left\{f_1,\cdots, f_n\right\}$.

\subsection{Central Controller Using RL}
\label{sec:RL-controller}

In addition, we design another central controller using deep reinforcement learning with our framework. The central controller acts as a feedback-loop controller system, which can be formulated as an MDP with the following definitions of key elements.

 \medskip \noindent \textbf{State.} In our scenario, the fast-forwarded videos of period $k$ from multiple agents are integrated into a single description, which is taken as the state $s_k$. To be more specific, we consider the concatenation of the average feature vector as a state, which is based on the fast-forwarded frames of different agents.  
    
 \medskip \noindent \textbf{Action.} At each control period $k$, we consider the action as the combination of different fast-forwarding strategies used in each agent. As we have $M$ available fast-forwarding strategies for all $N$ agents, i.e. $S = \left\{s^m, m \in [1,M]\right\}$, the entire action space $AS = \left\{a^1, a^2, ..., a^P\right\}$, where $P = M^N$. 
    
 \medskip \noindent \textbf{Reward.} After taking one action, i.e., selecting the proper fast-forwarding for each agent, the system transits from state $s_k$ to another state $s_{k+1}$ and an immediate reward $r_k = r(s_k, a_k, s_{k+1})$ is received by the system. The accumulated reward is further defined as 
    \begin{equation} \label{eqn:R}
        R = \sum_{k} \gamma^{k-1} r(s_k, a_k, s_{k+1}),
    \end{equation}
where $\gamma \in [0,1]$ is the discount factor for the rewards in the future. The goal of the central controller is to control the fast-forwarding paces of agents to maximize the coverage of the important scenes across multiple views and reduce the redundancy in the final summarized videos, by taking a sequence of actions. For a video available in the training set, we assume that the label of it is a binary vector, in which 1 indicates an important frame and 0 means an unimportant one.

After receiving the strategy instruction from the central controller, each agent will fast-forward its own video stream with the corresponding model and transmit the fast-forwarded video segment during the current control period. During a period of time $T$, an agent $n$ sends its fast-forwarded frames $b_n$ and the corresponding binary vector of selected frames $\hat{y}_n$ back to the central controller. With this information from all agents, the central controller receives the immediate reward at step k computed by the following equation:
\begin{equation} \label{eqn:reward}
r_k = \sum_{n=1}^{N} g(\mathbf{\hat{y}_{n,k}})^T g(\mathbf{y_{n,k}}) + \alpha \frac{\|g(\mathbf{\bar{y}_k})\|_1}{\sum_{n=1}^N \|\mathbf{\hat{y}_{n,k}}\|_1},
\end{equation}
where $\mathbf{\hat{y}_{n,k}}$ is the binary vector indicating selected frames from agent $n$ at time step $k$, and $\mathbf{y_{n,k}}$ is the ground truth binary vector of the view of agent $n$ during the current period. $\mathbf{\bar{y}_k}$ is the global ground truth binary vector of the scene at time step $k$, which is generated by 
\begin{equation} 
    \mathbf{\bar{y}_k} = min (\sum_{n=1}^N \mathbf{y_{n,k}}, \mathbf{1}),
\end{equation}
where the minimum is taken element-wise. 

The first term in Eqn.~\eqref{eqn:reward} gives higher rewards for the fast-forwarding action that selects the frames that match the ground truth better. As neighboring frames are often similar and share the same content, we hope to match the fast-forwarded result to the ground truth in a smoother fashion. That is, if the agent selects a frame that is close to the important frame, we will give some rewards, rather than no reward. To achieve this, we transfer the binary vector of selected frames for both the ground truth $\mathbf{y_{n,k}}$ and the transmitted results $\mathbf{\hat{y}_{n,k}}$ to a Gaussian distribution in a time window, denoted by the function $g(\cdot)$.
The second term in Eqn.~\eqref{eqn:reward} is used to reduce the redundancy in the fast-forwarded result. If the agents select more frames, the central controller will get a smaller reward for the current strategy selection.

 \medskip \noindent \textbf{Policy.} The policy $\pi$ decides the action to be executed at each time step by the system, i.e., it chooses the action for the system that maximizes the expected accumulated reward for the current step and the future as shown in Eqn.~\eqref{eqn:policy}. In other words, the policy finds the fast-forwarding strategy of each agent that gives a larger expected accumulated reward.
    \begin{equation} \label{eqn:policy}
        \pi(s_k) = arg \max_{a} E[R|s_k, a,\pi]
    \end{equation}

Similar to the training of FFNet, We utilize Q-learning to achieve this policy by evaluating the value of $E[R|s,a,\pi]$ as $Q(s,a)$ and use a feed-forward neural network to approximate the Q-value.

\begin{figure}[t]
	\begin{center}
		\includegraphics[width=0.9\linewidth]{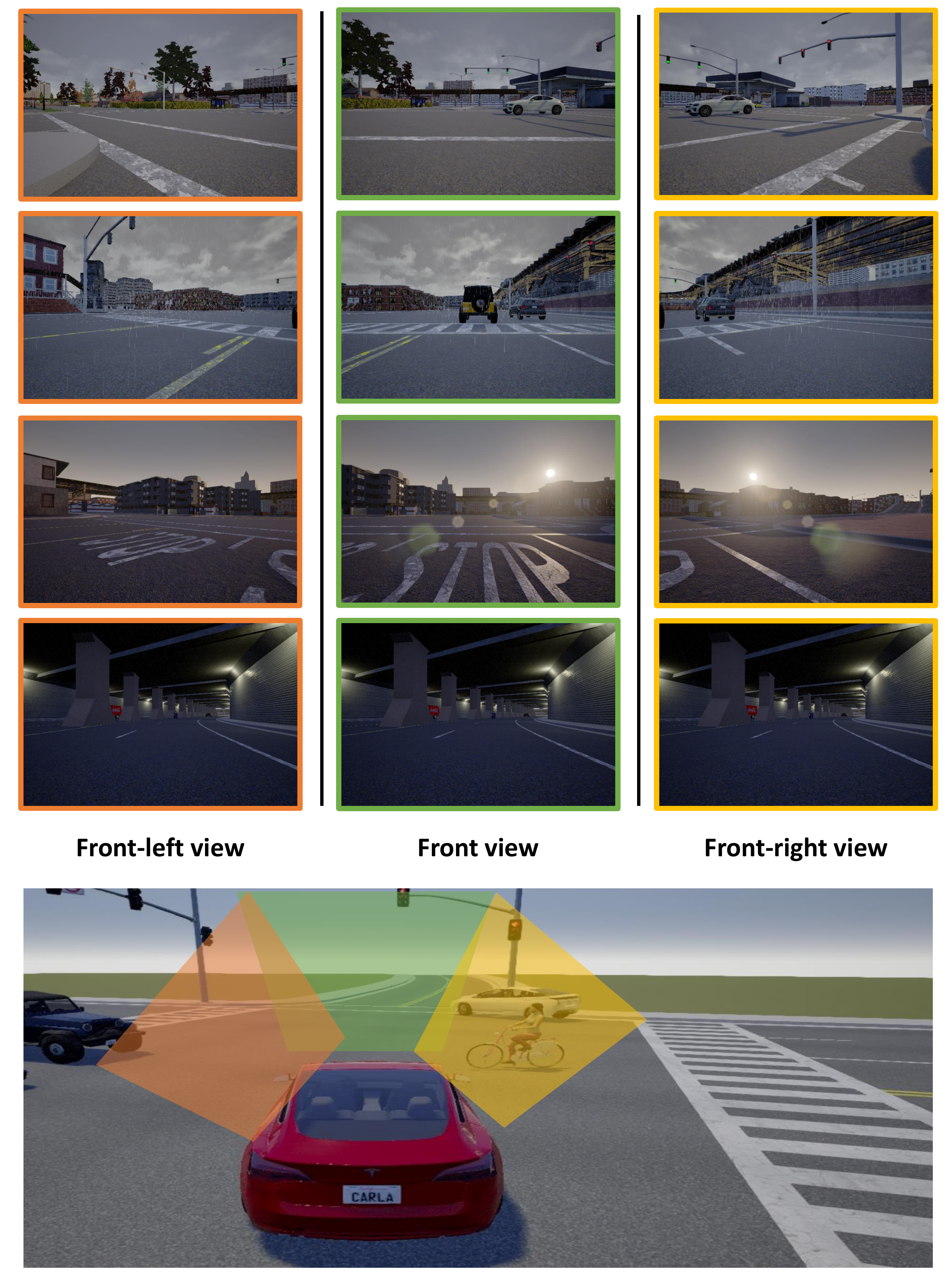}
	\end{center}
	\caption{\textbf{Some illustrative example frames from the CarlaSim dataset.} From left to right, the columns stand for frames from front-left, front, and front-right views. The CarlaSim dataset has multiple weather conditions, such as cloudy, rainy, and sunny (rows 1-3). Different terrains exist in the map, such as the tunnel in row 4. } 
	\label{dataset}
 \vspace{-12pt}
\end{figure}

\section{Experiments} \label{sec:exp}

In this section, we first present the experimental results of our MFFNet framework and its overall comparison with several single-agent fast-forwarding methods in the literature and FFNet, followed by its further comparison with FFNet in coverage-efficiency tradeoff and  high-redundancy cases. We then compare MFFNet with our previous distributed multi-agent fast-forwarding framework DMVF in detail.  
Finally, we also evaluate how communication issues may affect MFFNet, an important practical consideration. 

\subsection{Datasets}
We evaluate the performance of various methods on a multi-view video dataset VideoWeb~\cite{denina2011videoweb} with fixed cameras and on a self-built simulated multi-view dataset on moving platforms using the CARLA simulator~\cite{Dosovitskiy17}, referred to as CarlaSim.

\smallskip \noindent \textbf{VideoWeb.}
This dataset is captured in a realistic multi-camera network environment that involves multiple persons performing many different repetitive and non-repetitive activities. Same as in~\cite{lan2020distributed}, we use the Day 4 subset of the VideoWeb dataset, which contains multiple vehicles and persons. It has 6 scenes and each scene has 6 views of videos. All videos are captured at $640 \times 480$ resolution and approximately 30 frames/second. The dataset includes the labels for important activities, based on which, we can generate a binary indicator for each frame to label its importance. That is, if a frame contains the labeled important activities, it will be labeled as an important frame with the binary indicator as 1 (otherwise, as 0). With such a frame importance indicator, we can generate a global ground truth across views for evaluation purpose.

\begin{table}[t]
\begin{center}
\begin{tabular}{|c|c|}
\hline
Specification & Value \\
\hline
City & Town-3\\
\hline
Number of videos & 18 \\
\hline
Video length & 10000 \\
\hline 
Resolution & 720 x 480 \\
\hline
Camera & front, front-left, front-right \\
\hline
Terrain & 5-lane junction, roundabout, unevenness, tunnel\\
\hline
Weather & dynamic cloudiness, precipitation, sun angle \\
\hline
\end{tabular}
\end{center}
\caption{\textbf{Specifications in CARLA for generating Carlasim dataset.}}
\label{tab:spec}
\vspace{-12pt}
\end{table}

\smallskip \noindent \textbf{CarlaSim.}
 CARLA is a simulator for urban autonomous driving. It provides open digital assets (urban layouts, buildings, and vehicles) and supports flexible specifications of sensor suites, environmental conditions, full control of all static and dynamic actors, map generation, and more. For generating the CarlaSim dataset, we utilize the Town3 environment in the CARLA simulator, which has a 5-lane junction, a roundabout, unevenness, a tunnel, and so on. The multi-view videos are captured by putting multiple cameras on an autonomous car, which runs with a built-in autonomous driving controller. Detailed specifications for generating the video data are shown in Tab.~\ref{tab:spec}. We generate a binary indicator for each video frame according to the existence of vehicles in the view. If a frame captures a nearby vehicle (i.e., with size $>$ 150 pixels), it will be labeled as an important frame. The global ground truth for evaluation is generated by the same method as for the VideoWeb dataset. Fig.~\ref{dataset} shows some example frames in the CarlaSim dataset. From left to right, the columns stand for frames from front-left, front, and front-right views. As the data is collected on a moving platform, it captures more dynamic scenarios than the existing datasets that use stationary cameras (such as VideoWeb) and can help validate the efficacy of our methods in those dynamic scenarios.

\subsection{Experimental Setup}

\smallskip \noindent \textbf{Implementation Details.}
Our MFFNet is implemented using the TensorFlow library. The fast-forwarding agents are all modeled as 4-layer neural networks. $\epsilon$-greedy strategy is used to better explore the state space during the training process. In the following experiments, we explore the scenarios of both 3 views ($N = 3$) and 6 views ($N = 6$), and set the similarity threshold  to $\rho = 0.525$ and $\rho = 0.575$, respectively. The strategy computation threshold $\tau$ is set to 0.4. The strategy update period $T$ is set to 100 frames of the raw video inputs.  The 3 strategies used in our framework are FFNet and its variants as defined in Sec. \ref{sec:agent}. The operating points of agents with the slow, normal and fast strategies are shown in Tab.~\ref{tab:per_stra} for VideoWeb and Tab.~\ref{tab:carla_ffnet} for CarlaSim. 

\begin{table}[t]
\begin{center}
\begin{tabular}{|c|c|c|c|}
\hline
Strategy & Slow & Normal & Fast \\
\hline
Processing rate(\%) & 8.69 & 6.02 & 3.73\\
\hline
3-view Coverage(\%) & 66.22 & 52.88 & 48.38 \\
\hline
6-view Coverage(\%) & 73.45 & 61.91 & 55.89 \\
\hline
\end{tabular}
\end{center}
\caption{\textbf{Operating points of strategies on VideoWeb.}}
\label{tab:per_stra}
\vspace{-6pt}
\end{table}

\begin{table}[t]
\begin{center}
\begin{tabular}{|c|c|c|c|}
\hline
Method & Slow & Normal & Fast \\
\hline
coverage(\%) & 80.78 & 67.83 & 60.78  \\
\hline
Processing Rate(\%) & 18.06 & 14.76 & 7.09  \\
\hline
\end{tabular}
\end{center}
\caption{\textbf{Operating points of strategies on CarlaSim.} }
\label{tab:carla_ffnet}
\vspace{-12pt}
\end{table}

Each video frame is represented by the penultimate layer (pool 5) of the GoogLeNet model~\cite{szegedy2015going} (1024-dimensions). Each baseline algorithm is evaluated with the same neighboring window extension as ours. We randomly use 80\% of the videos for training and the remaining 20\% for testing. We report the average performance on 5 rounds of experiments.

\begin{table*}[t]
\begin{center}
\begin{tabular}{|c|c|c|c|c|c|c|c|}
\hline
Methods & Random & Uniform & OK & SC & SMRS  & FFNet & MFFNet\\
\hline
VideoWeb 3-view Coverage (\%) & 41.33 & 27.79 & 39.92 & 42.10 & 31.10  & 52.88 & \textbf{53.66}\\
\hline
VideoWeb 3-view Processing rate (\%) & 4.40 & 4.00 & 100 & 100 & 100  & 6.02 & \textbf{5.46} \\
\hline
VideoWeb 6-view Coverage (\%) & 50.78 & 25.80 & 50.21 & 44.74 & 42.36  & 61.91 & \textbf{61.92} \\
\hline
VideoWeb 6-view Processing rate (\%) & 4.20 & 3.70 & 100 & 100 & 100  & 6.02 & \textbf{5.58}\\
\hline
CarlaSim Coverage (\%) & 55.69 & 36.74 & 52.24 & 51.80 & 46.85 &  67.83 & \textbf{68.65}\\
\hline
CarlaSim Processing rate(\%) & 6.50  & 5.40  & 100  & 100 & 100 &  14.76 & \textbf{12.96}\\
\hline
\end{tabular}
\end{center}
\caption{\textbf{Comparison of MFFNet with single-agent fast-forwarding approaches for both VideoWeb and CarlaSim datasets.}}
\label{tab:main}
\vspace{-6pt}
\end{table*}

\begin{figure*}[t]
	\begin{center}
		\includegraphics[width=\linewidth]{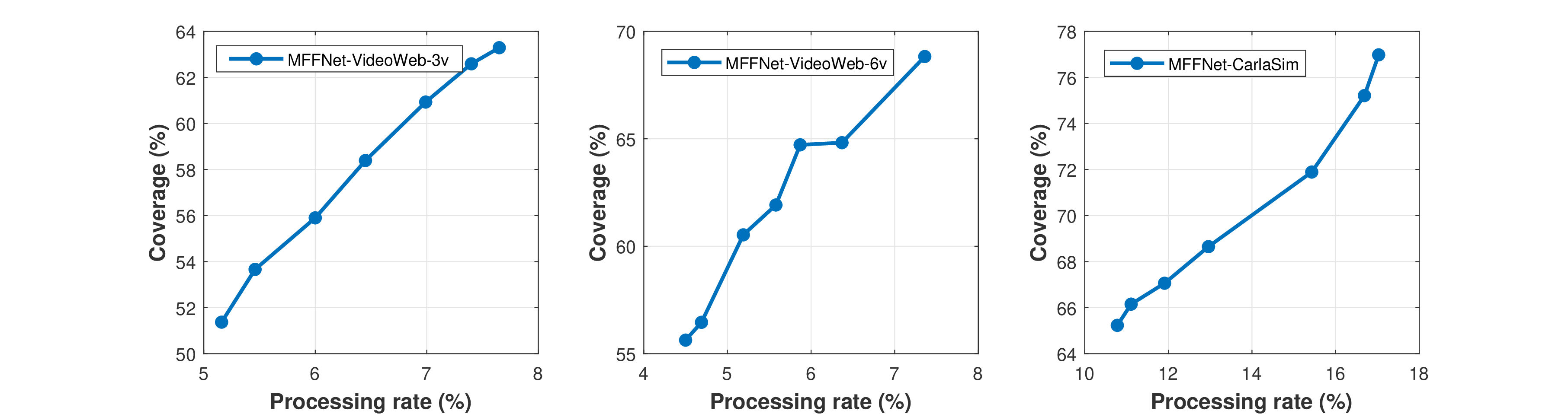}
	\end{center}
	\vspace{-6pt}
	\caption{\textbf{Trade-off between coverage and processing rate in MFFNet for 3-view and 6-view scenarios of VideoWeb dataset.} By tuning the similarity threshold $\rho$ (marked in the figure), different levels of tradeoff can be achieved.}  
	\label{tradeoff}
 \vspace{-6pt}
\end{figure*}

\smallskip \noindent \textbf{Evaluation Metrics.} We evaluate the performance of methods with a \emph{coverage} metric and a \emph{processing rate} metric. The coverage metric evaluates how well the resulting fast-forwarding videos across multiple agents cover the important frames in the ground truth. It is computed as the percentage of the important frames that are included in the fast-forwarding videos across agents. In other words, if an important frame is included in any one of the agents' fast-forwarding videos, it will be considered as covered. The processing rate metric measures the percentage of the frames being processed by the system.

\smallskip \noindent \textbf{Comparison Methods.}
We compare our MFFNet with the following methods for video fast-forwarding and video summarization: 
(1) \textbf{Random}, which skips the incoming frames randomly. 
(2) \textbf{Uniform}, which fast-forwards the video uniformly.
(3) \textbf{Online Kmeans (OK)}~\cite{arthur2007k}, a clustering-based method working in an online update fashion. The summary result consists of the frames that are the closest to the centroid in each cluster.
(4) \textbf{Spectral Clustering (SC)}~\cite{von2007tutorial}, a clustering-based method that provides several clusters from all the frames in a video. The summary is composed by the frames that are closest to each centroid. 
(5) \textbf{Sparse Modeling Representative Selection (SMRS)}~\cite{elhamifar2012see}, which takes the entire video as the dictionary and finds the representative frames based on the zero patterns of the sparse coding vector.
(6) \textbf{FFNet}~\cite{lan2018ffnet}, the method we developed for single-agent video fast-forwarding. 
(7) \textbf{DMVF}~\cite{lan2020distributed}, the distributed multi-agent fast-forwarding method we developed.

\subsection{Comparison of MFFNet with Single-agent Fast-forwarding Approaches}

Tab.~\ref{tab:main} shows the coverage metric and the processing rate of the single-agent fast-forwarding approaches in the literature, FFNet, and MFFNet,  
on the VideoWeb dataset for the 3-view and 6-view scenarios and the CarlaSim dataset. Note that in the cases of single-agent approaches (including FFNet), every view/camera uses the same approach and configuration. In contrast, a multi-agent approach like MFFNet coordinates the operations of multiple views. From the table, we can clearly see its advantage. 
More specifically: 
\begin{myitemize}
    \item For those methods that require processing the entire video (processing rate of 100\%), i.e., OK, SC and SMRS, our approach MFFNet achieves higher coverage (more than 25\% increase) and much lower processing rate.
    \item When compared with Random and Uniform methods, MFFNet offers significant improvement in coverage with a modest increase in processing rate.
    \item When compared with FFNet, our state-of-the-art single-agent approach, MFFNet achieves a slightly better coverage while reducing the processing rate by $9.3\%$ in VideoWeb 3-view, $7.3\%$ in VideoWeb 6-view and $12.20\%$ in CarlaSim. This shows that MFFNet is able to further reduce the computation load in the fast-forwarding process while offering the same (or higher) level of coverage of important frames. 
\end{myitemize}

\subsection{Further Comparison of MFFNet with FFNet}

\medskip \noindent \textbf{Enabling Flexible Coverage-Efficiency Tradeoff.} When deploying a video fast-forwarding strategy, the goal of achieving high efficiency (i.e., low processing rate) contradicts with the goal of maintaining high coverage, and the designers may want to trade off between the two metrics. To enable such tradeoff, MFFNet incorporates a tunable parameter, i.e., the similarity threshold $\rho$. Fig.~\ref{tradeoff} shows that by changing $\rho$, different levels of tradeoff between coverage and efficiency can be easily achieved on VideoWeb dataset for 3-view and 6-view scenarios and on CarlaSim. This is much more flexible and systematic than simply deploying FFNet on each agent and manually trying their skipping speeds.

\smallskip \noindent \textbf{Addressing High-redundancy Cases.} 
The different views in the VideoWeb dataset have a modest level of redundancy across them. When the redundancy level is higher, the improvement of our MFFNet over FFNet will be even more significant.
Here we consider the extreme case where each view has the same video data, i.e., the highest level of redundancy. The fast-forwarding performance of MFFNet and FFNet in both VideoWeb and CarlaSim is shown in Tab.~\ref{tab:extreme}. As CarlaSim only has 3 different camera views, thus no results are available for MFFNet-6v. Note that FFNet does not have any strategy changes in different settings, so its results for 3-view and 6-view are the same for the extreme case.
From the result, we can see that MFFNet can achieve much higher coverage and lower processing rate than FFNet.
 
 \begin{table}[t]
\begin{center}
\begin{tabular}{|c|c|c|c|}
\hline
Methods & FFNet & MFFNet-3v & MFFNet-6v \\
\hline
VideoWeb Coverage(\%) & 54.10 & 71.93 & 75.61 \\
\hline
VideoWeb Processing rate(\%) & 8.69 & 5.30 & 4.53 \\
\hline
CarlaSim Coverage(\%) & 52.38 & 79.31 & / \\
\hline
CarlaSim Processing rate(\%) & 14.76 & 8.09 & / \\
\hline
\end{tabular}
\end{center}
\caption{\textbf{Comparison of MFFNet and FFNet in the extreme case, where all views have the same data.} }
\label{tab:extreme}
\vspace{-6pt}
\end{table}

\subsection{Comparison of MFFNet with Distributed Multi-agent Framework DMVF}

In this section, we compare MFFNet with our distributed multi-agent video fast-forwarding framework DMVF~\cite{lan2020distributed}, on both VideoWeb 6-view and CarlaSim datasets. The results are shown in Tab.~\ref{tab:dmvf}. We have the following findings:
\begin{myitemize}
    \item MFFNet and DMVF are comparable in coverage and processing rates on VideoWeb and CarlaSim.  On VideoWeb, DMVF achieves better coverage while MFFNet achieves better coverage on CarlaSim.
    \item On both datasets, MFFNet has less communication load (-44\% for VideoWeb and -15\% for CarlaSim) and higher frame rate (+34\% on VideoWeb and +93\% on CarlaSim). This is because that DMVF is a distributed method. The same information from one agent may need to be sent multiple times and the framework needs to reach a consensus on the strategy update, which leads to a higher communication load and longer communication delay.
\end{myitemize}

While MFFNet has the advantages on less communication load and higher frame rate, DMVF is more flexible to utilize as it does not need a centralized infrastructure and the connections among agents can be adjusted according to system needs and agent capabilities. Both centralized and distributed methods could be suitable for improving the efficiency of a network of resource-limited agents with cameras, which can be used in tasks such as search and rescue, wide-area surveillance, and environment monitoring. Considering the advantages of each method, the choice between them depends on the practical application scenario. If we have a stable centralized infrastructure and each agent is able to reliably connect to the central controller, the centralized MFFNet might be a better choice as it can further reduce the communication load and improve the overall efficiency. However, in some cases (e.g., in an adversarial environment) we do not have a stable and capable centralized infrastructure, and some agents may not be able to reliably connect to the central node due to their physical distance or own resource limitations, in which case DMVF might be a better choice. 

 \begin{table*}[tb]
\begin{center}
\begin{tabular}{|c|c|c|c|c|}
\hline
Method-Dataset & DMVF-VideoWeb & MFFNet-VideoWeb & DMVF-CarlaSim & MFFNet-CarlaSim \\
\hline
coverage(\%) & 65.87 & 61.92 & 64.54 & 68.65 \\
\hline
Processing Rate(\%) & 5.06  & 5.58 & 12.33 & 12.96 \\
\hline
Communication p2p(GB)  & 0.18 & / & 0.20 & / \\
\hline
Communication central (GB) & / & 0.10 & / & 0.17\\
\hline
Total Communication (GB) & 0.18 & 0.10 & 0.20 & 0.17 \\
\hline
Summary to Server (GB) & 3.59 & 3.22 & 2.27  & 2.46\\
\hline
FPS & 313 & 419 & 119 & 230 \\
\hline
\end{tabular}
\end{center}
\caption{\textbf{Comparison of MFFNet with distributed multi-agent fast-forwarding framework DMVF.} }
\label{tab:dmvf}
\vspace{-6pt}
\end{table*}

\subsection{Impact of Communication on MFFNet}

For a multi-agent strategy such as MFFNet, communication issues such as desynchronization or packet losses could have a major impact in practice, especially in the case of wireless communication (in~\cite{lan2020distributed}, the impact of network connectivity on DMVF was studied). In this section, we evaluate the performance of MFFNet under the impact of such communication issues, using the VideoWeb dataset for illustration.

\begin{figure}[t]
\centering
\includegraphics[width=0.95\linewidth]{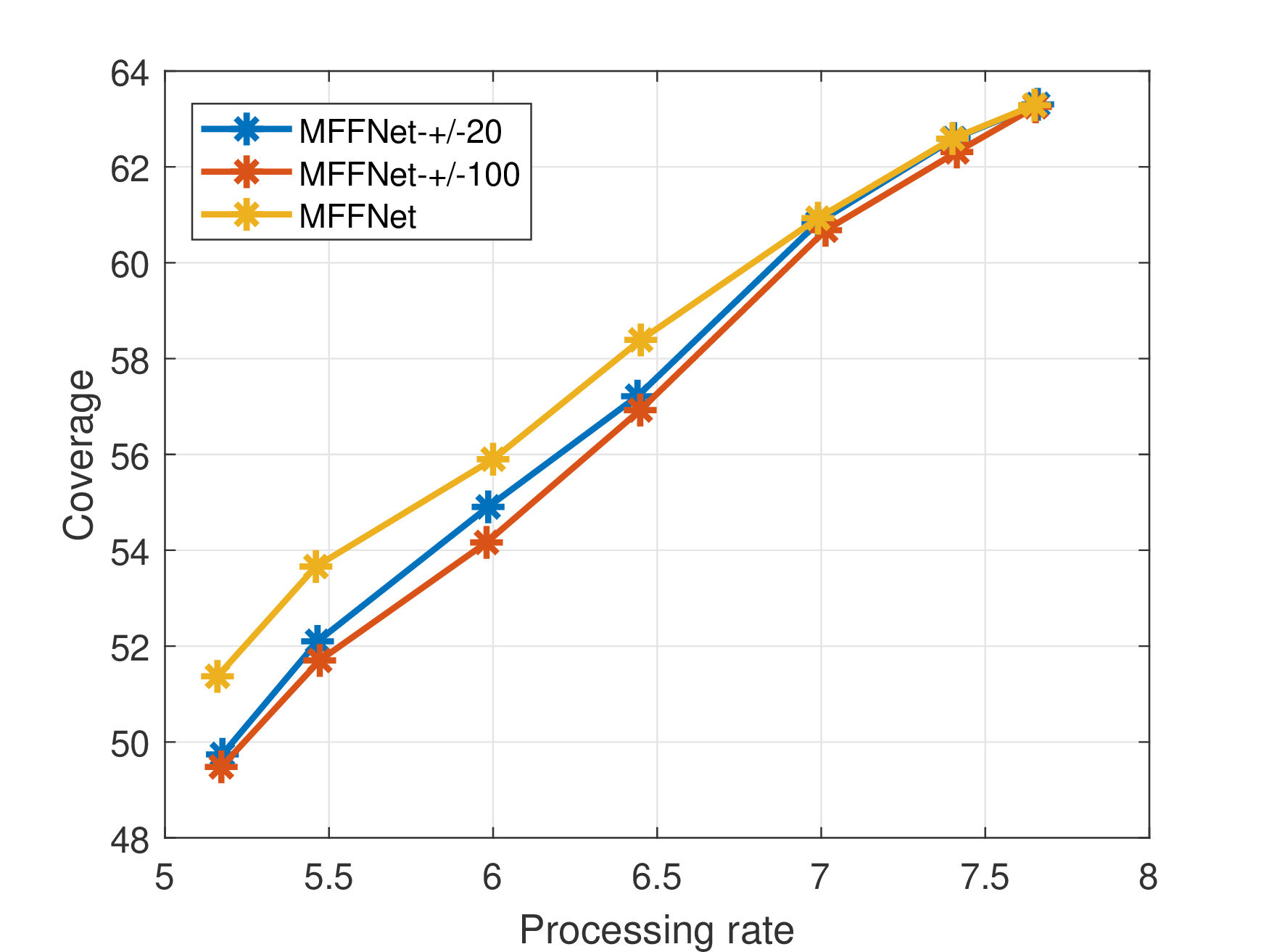}
\caption{\textbf{Effect of desynchronization on MFFNet in 3-view VideoWeb scenario.} The desynchronization has some effect on the coverage of MFFNet, but the drop is not too significant. } 
\label{fig:asyn_delay} 
\end{figure}

\smallskip \noindent \textbf{Desynchronization.} In this experiment, we consider the desynchronization of one view with respect to the others. For instance, frame 20 from one view may be taken physically at the same time as frame 0 of other views, but is given a time tag that is the same as frame 20 of other views (this could happen due to the desynchronization of camera clocks). Fig.~\ref{fig:asyn_delay} shows the results on the 3-view scenario when one view is 20 or 100 frames desynchronized (either ahead or behind) with the other views. We can see that the desynchronization indeed has some effect on MFFNet coverage, but the drop is not too significant. Similar results can be observed for the 6-view scenario. In practice, with a decent clock synchronization scheme, we should be able to maintain the desynchronization to be under 20 frames.

\medskip \noindent \textbf{Packet Losses.} We consider the cases where a packet from an agent to the central controller may be lost due to communication disturbance. Each packet is the fast-forwarded segments from 100-frame raw videos at an agent. Tab.~\ref{tab:pack_loss} shows the coverage of MFFNet when the packet loss probability varies from $2.5\%$ to $10\%$. 
We can see that the drop is not very significant. Moreover, most of the coverage drop is due to the loss of data itself rather than the strategy selection process.
\begin{table}
\begin{center}
\begin{tabular}{|c|c|c|c|c|}
\hline
Loss probability & 2.5\% & 5.0\% & 7.5\% & 10.0\% \\
\hline
3-view coverage(\%) & 52.00 & 50.98 & 49.43 & 49.00 \\
\hline
6-view coverage(\%) & 60.43 & 60.08 & 59.90 & 57.89 \\
\hline
\end{tabular}
\end{center}
\caption{\textbf{Effect of packet losses on MFFNet in 3-view and 6-view scenarios in VideoWeb.} The performance is slightly affected by the packet loss (less than 10\% degradation in 10\% loss probability).}
\label{tab:pack_loss}
\vspace{-6pt}
\end{table}

\subsection{Study of Central Controller Designs in MFFNet}

The above results of MFFNet are based on the heuristic central controller design introduced in Sec.~\ref{sec:controller}, with explicit similarity computation and strategy computation. 
In this section, we compare such central controller design with the RL-based design introduced in Sec.~\ref{sec:RL-controller}, using the VideoWeb 3-view case as an example. The results are shown in Tab.~\ref{tab:dqn}, where MFFNet is the heuristic central controller based on similarity computation, and 
MFFNet-DQN-0 and MFFNet-DQN-1 represent two RL-based central controllers using DQN models with the trade-off factor $\alpha$ in Eqn.~\eqref{eqn:reward} set to $0$ and $1$, respectively. From the table, we have the following observations: 1) The RL-based central controllers using DQN have higher coverage than the heuristic one based on similarity computation but also have much higher processing rate. 2) Setting the trade-off term $\alpha$ in the immediate reward to 1 can help lower the processing rate but also degrade the coverage. Note that the choice of which central controller to use depends on the trade-off preference on coverage or processing rate. 

\begin{table}[t]
\begin{center}
\begin{tabular}{|c|c|c|c|}
\hline
Method & MFFNet & MFFNet-DQN-0 & MFFNet-DQN-1\\
\hline
Coverage(\%) & 53.66 & 64.80 & 64.24\\
\hline
Processing Rate(\%) & 5.46 & 7.32 & 7.07\\
\hline
\end{tabular}
\end{center}
\caption{\textbf{Comparison of different controllers in MFFNet.}  }
\label{tab:dqn}
\vspace{-12pt}
\end{table}

\subsection{Deployment of MFFNet on Embedded Platform} 

We deployed MFFNet on an actual embedded platform to evaluate its efficiency. The central controller is implemented on a Dell Precision 5820 Tower workstation with a 3.6 GHz Xeon W-2123 CPU and 16GB memory, and the agents are run on Nvidia Jetson TX2. The communication between the central controller and the agents is implemented with a wireless network using TCP. For MFFNet, the average frame rate is 661 FPS for the 3-view scenario and 419 FPS for the 6-view scenario (note that only a fraction of these frames will be actually processed), showing its capability to work efficiently and effectively with real-time speed on embedded processors. 

\section{Conclusion}
In this paper, we first summarize our previous work on the single-agent video fast-forwarding method FFNet and distributed multi-agent video fast-forwarding framework DMVF, and then present a new centralized multi-agent fast-forwarding
framework MFFNet. The MFFNet framework includes
a set of multi-strategy fast-forwarding agents that can
adapt to different fast-forwarding paces, and a central controller
that can choose the proper pace for every agent and
generate a compact summary of the scene. We conducted a series of experiments on a real-world surveillance video dataset and a new simulated driving dataset, for MFFNet, DMVF, FFNet, and several methods in the literature. Experimental results demonstrate that our two collaborative multi-agent video fast-forwarding approaches, MFFNet and DMVF, can achieve better scene coverage and lower frame processing rate than applying single-agent fast-forwarding approaches on multiple agents without coordination. The experiments also demonstrate the trade-off between MFFNet and DMVF, the impact of communication disturbance, and the choice of different central controller designs.





\section*{Acknowledgment}
We gratefully acknowledge the support from NSF grants 1834701, 1724341, 2038853, 2024774, and ONR grant N00014-19-1-2496.

\ifCLASSOPTIONcaptionsoff
  \newpage
\fi



%


\bibliographystyle{IEEEtran}
\bibliography{shuyue_bib}

%


\begin{IEEEbiography}
    [{\includegraphics[width=1in,height=1.2in,clip,keepaspectratio]{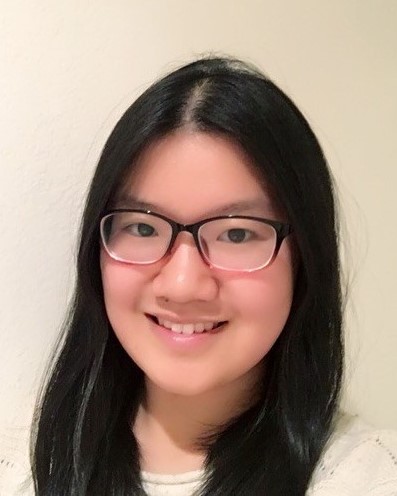}}]{Shuyue Lan}
graduated from Northwestern University with a Ph.D. in Computer Engineering in 2021. She spent her first two years of Ph.D. study at UC Riverside. Previously, she received her Bachelor’s degree in Automation from University of Science and Technology of China (USTC) in 2015. Her research interest includes Computer Vision, Machine Learning and Cyber-physical Systems. Currently, her work is focusing on high-performance deep learning inference workflow.
\end{IEEEbiography}

\begin{IEEEbiography}
    [{\includegraphics[width=1in,height=1.25in,clip,keepaspectratio]{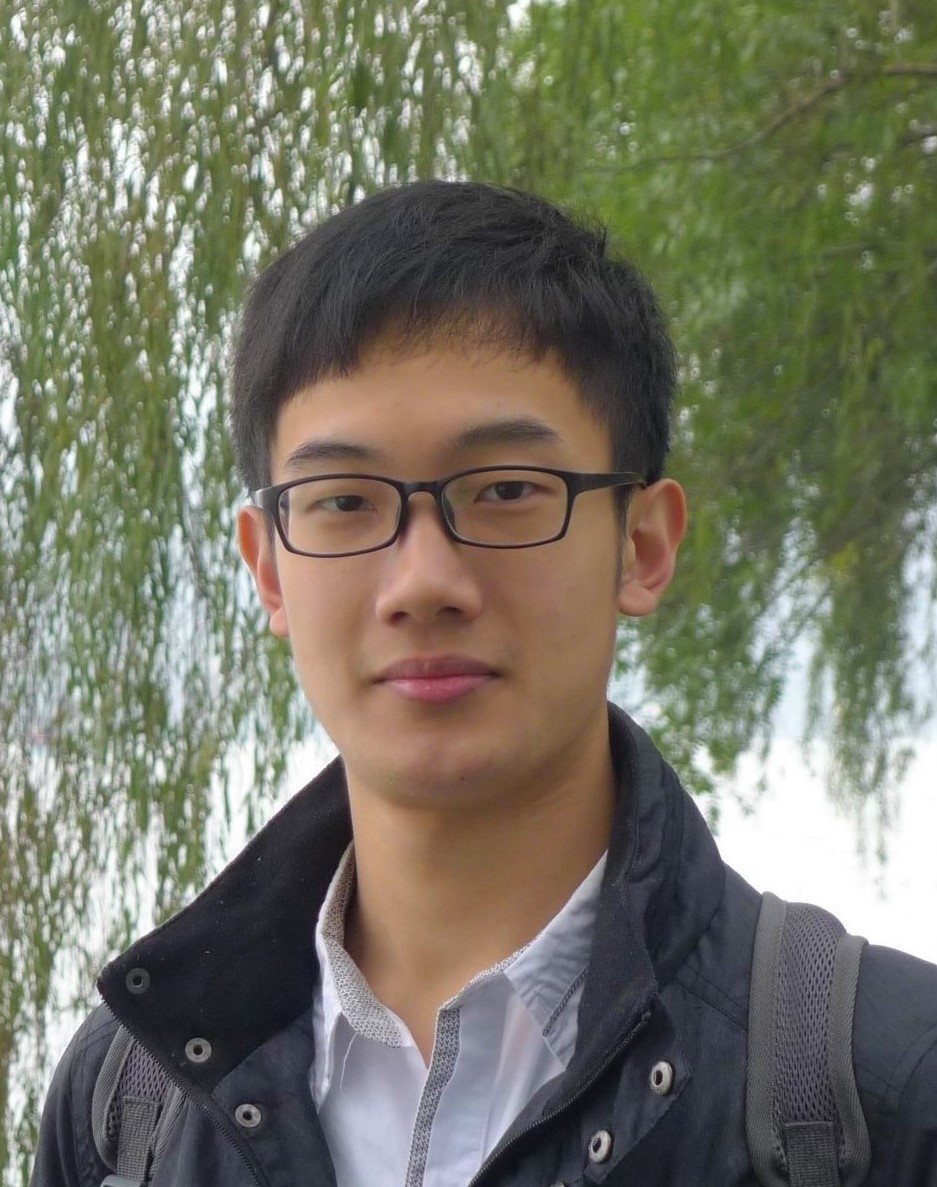}}]{Zhilu Wang}
graduated from Northwestern University with a Ph.D. degree in Computer Engineering in 2022. He began his doctoral career at University of California, Riverside in 2016. Prior to that, he received his B.S. degree in Applied Physics from University of Science and Technology of China in 2016. He received the Best Paper Award at the 2022 ACM/IEEE DATE conference and the Best Thesis Award in Computer Engineering from Northwestern University. His research interest includes formal verification, machine learning, real-time systems, and cyber-physical systems.
\end{IEEEbiography}

\begin{IEEEbiography}[{\includegraphics[width=1in,height=1.25in,clip,keepaspectratio]{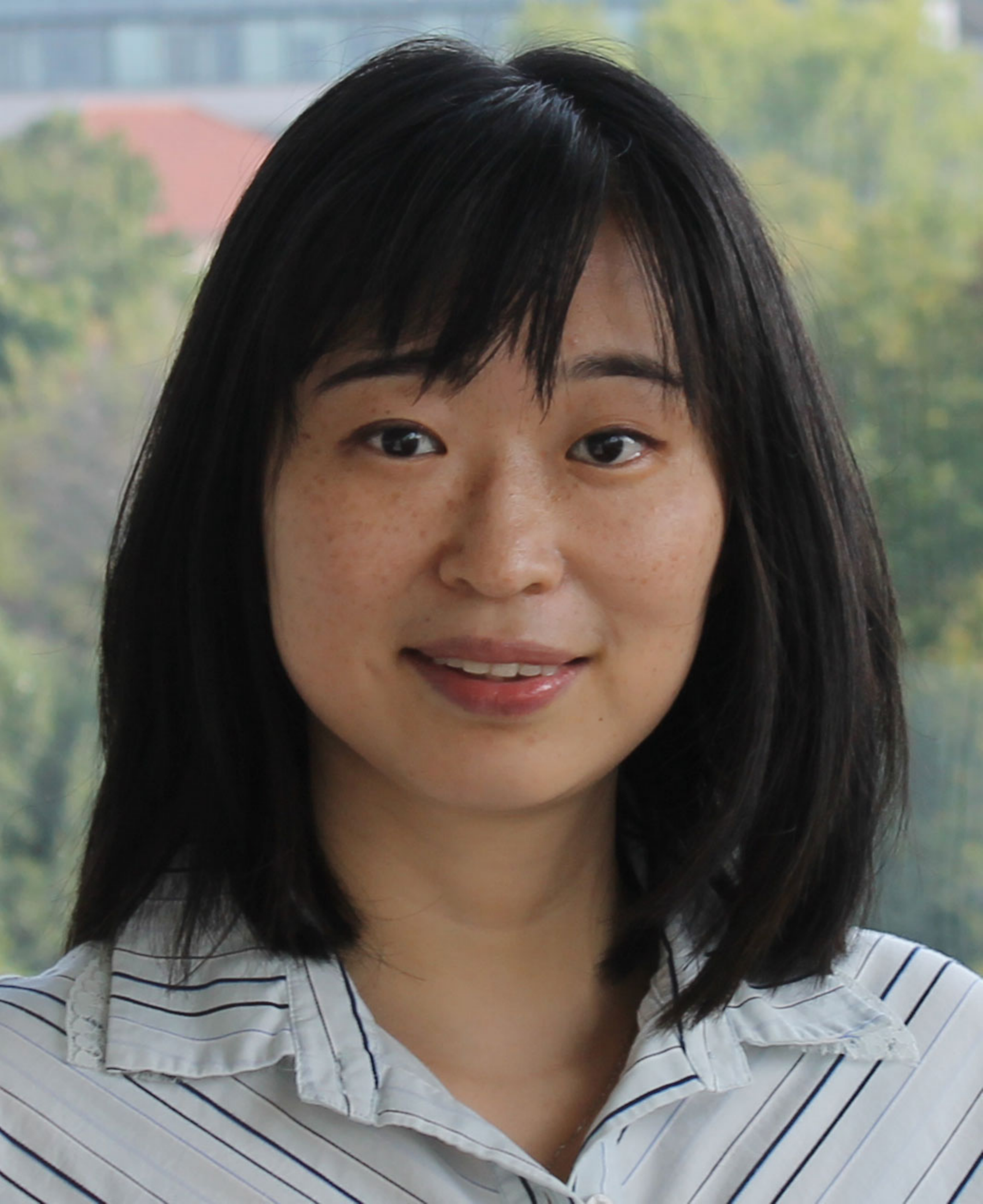}}]{Ermin Wei} is currently an Assistant Professor at the Electrical and Computer Engineering Department and Industrial Engineering and Management Sciences Department of Northwestern University. She completed her PhD studies in Electrical Engineering and Computer Science at MIT in 2014, advised by Professor Asu Ozdaglar, where she also obtained her M.S.. She received her undergraduate triple degree in Computer Engineering, Finance and Mathematics with a minor in German, from University of Maryland, College Park. Wei has received many awards, including the Graduate Women of Excellence Award, second place prize in Ernst A. Guillemen Thesis Award and Alpha Lambda Delta National Academic Honor Society Betty Jo Budson Fellowship. Her team also won the 2nd place in the Grid Optimization (GO) competition 2019, an electricity grid optimization competition organized by Department of Energy. Wei's research interests include distributed optimization methods, convex optimization and analysis, smart grid, communication systems and energy networks and market economic analysis.
\end{IEEEbiography}

\begin{IEEEbiography}
    [{\includegraphics[width=1in,height=1.25in,clip,keepaspectratio]{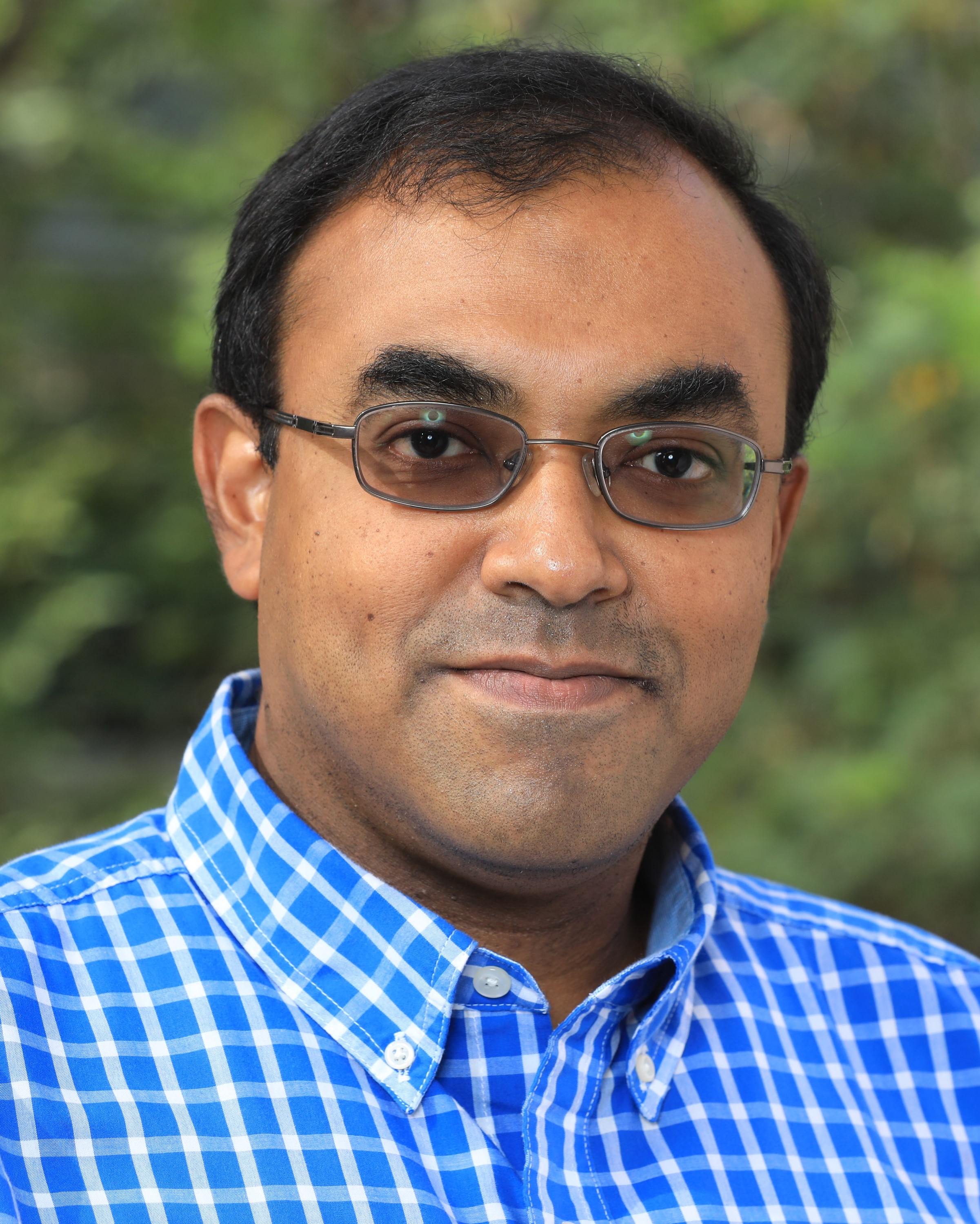}}]{Amit K. Roy-Chowdhury} received his PhD from the University of Maryland, College Park (UMCP) in 2002 and joined the University of California, Riverside (UCR) in 2004 where he is a Professor and Bourns Family Faculty Fellow of Electrical and Computer Engineering, Director of the Center for Robotics and Intelligent Systems, and Cooperating Faculty in the department of Computer Science and Engineering. He leads the Video Computing Group at UCR, working on foundational principles of computer vision, image processing, and statistical learning, with applications in cyber-physical, autonomous and intelligent systems. He has published over 200 papers in peer-reviewed journals and conferences. He has published two monographs: Camera Networks: The Acquisition and Analysis of Videos Over Wide Areas and Person Re-identification with Limited Supervision. He is on the editorial boards of major journals and program committees of the main conferences in his area. His students have been first authors on multiple papers that received Best Paper Awards at major international conferences. He is a Fellow of the IEEE and IAPR, received the Doctoral Dissertation Advising/Mentoring Award 2019 from UCR, and the ECE Distinguished Alumni Award from UMCP.
\end{IEEEbiography}

\begin{IEEEbiography}[{\includegraphics[width=1in,height=1.25in,clip,keepaspectratio]{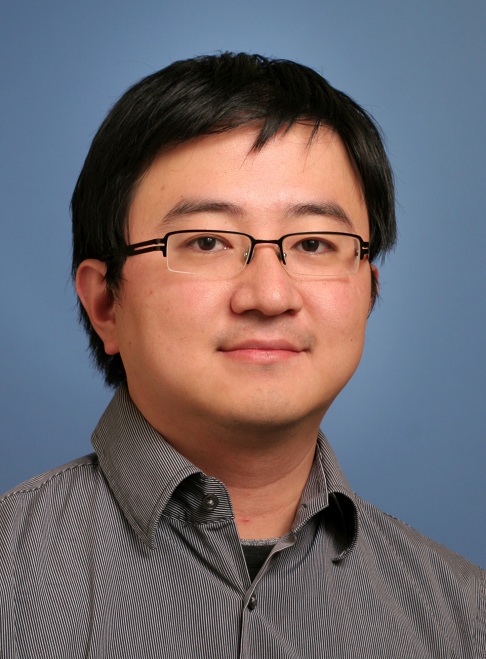}}]{Qi Zhu} is an Associate Professor at the ECE Department in Northwestern University. He received a Ph.D. in EECS from University of California, Berkeley in 2008, and a B.E. in CS from Tsinghua University in 2003. His research interests include design automation for cyber-physical systems (CPS) and Internet of Things, safe and secure machine learning for CPS and IoT, cyber-physical security, and system-on-chip design, with applications in domains such as connected and autonomous vehicles, energy-efficient smart buildings, and robotic systems. He is a recipient of the NSF CAREER award, the IEEE TCCPS Early-Career Award, and the Humboldt Research Fellowship for Experienced Researchers. He received best paper awards at DAC 2006, DAC 2007, ICCPS 2013, ACM TODAES 2016, and DATE 2022. He is the Conference Chair of IEEE TCCPS, and Young Professionals Coordinator at IEEE CEDA. He is an Associate Editor for IEEE TCAD, ACM TCPS, and IET Cyber-Physical Systems: Theory \& Applications, and has served as a Guest Editor for the Proceedings of the IEEE, ACM TCPS, IEEE T-ASE, Elsevier JSA, and Elsevier Integration, the VLSI journal.
\end{IEEEbiography}







\end{document}